\documentclass[12pt]{article}
\usepackage{amsthm,amsmath}
\usepackage[utf8]{inputenc} 
\usepackage{amsmath}
\usepackage{marginnote}
\usepackage{amsmath, bm, setspace}
\usepackage{tabularx,colortbl}
\usepackage{float}
\usepackage{caption}
\usepackage{booktabs,multirow,array}
\usepackage{comment}
\usepackage{chngpage}
\usepackage{subcaption}
\usepackage{csquotes}
\usepackage{array}
\usepackage{indentfirst}
\usepackage{graphicx, amsmath, bm, rotating, pdflscape, xcolor, longtable}
\usepackage{float}
\usepackage{color}
\usepackage{amssymb}
\usepackage{dcolumn}
\usepackage{array}
\usepackage{siunitx}
\usepackage{listings}
\usepackage{etoolbox}
\usepackage{cite}
\usepackage{endnotes} 
\usepackage{authblk}
\usepackage{hanging}
\usepackage[ruled,linesnumbered,noresetcount]{algorithm2e}

\begin{document}
\baselineskip 12pt

\begin{center}
\textbf{\large Metaheuristics is All You Need 
} \\

\vspace{1.5cc}
{ \sc Elvis Han Cui$^{1}$, Haowen Xu$^{2}$ and Weng Kee Wong$^{3,*}$}\\

\vspace{0.3 cm}

{\small $^{1}$Department of Neurology,
University of California, Irvine, cuieh@hs.uci.edu\\
$^{2}$Department of Biostatistics,
University of California, Los
Angeles, haowenxu@g.ucla.edu\\
$^{3}$Department of Biostatistics,
University of California, Los
Angeles, wkwong@g.ucla.edu\\
$^{*}$ Correspondence author.}%
 \end{center}
\vspace{1.5cc}

\small
\begin{abstract}
  \noindent  Optimization plays an important role in tackling public health problems. Animal instincts can be used  effectively to solve complex public health management issues by providing optimal or approximately optimal solutions to complicated  optimization problems common in public health. BAT algorithm is an exemplary member of a class of nature-inspired metaheuristic optimization algorithms  and designed to outperform  existing metaheuristic algorithms in terms of efficiency and accuracy. It's inspiration comes from the foraging behavior of group of microbats that use echolocation to find their target in the surrounding environment. In recent years, BAT algorithm has been extensively used by researchers in the area of optimization, and various variants of BAT algorithm have been developed to improve its performance and extend its application to diverse disciplines. This paper first reviews the basic BAT algorithm and its variants, including their applications in various fields. As a specific application, we apply the BAT algorithm to a biostatistical estimation problem and show it has some clear advantages over existing algorithms.

\vspace{0.95cc}
\parbox{24cc}{{\it Key words and Phrases}: 
Bat algorithm, maximum likelihood estimates, relative risk, variants
}
\end{abstract}

\section{Introduction}

Optimization algorithms are widely used in many fields, such as engineering, finance, computer science and statistics. As the technology advances, we have sophisticated tools that could help us in formulating and solving increasingly difficult and high-dimensional optimization problems. However, resources such as computing power and time can still be limited especially when we want to find solutions to problems quickly. Therefore, more flexible, powerful and more efficient optimization algorithms are constantly needed in solving problems with high complexity.

Metaheuristics is a class of optimization algorithms designed to solve complicated problems with improved efficiency. This type of optimization algorithm is also widely used in solving recent public health problems. For example, \cite{shindi2020combined} proposed two hybrid methodologies to solve a constrained multi-objective optimization problem in cancer chemotherapy. \cite{cui2022single} applied Metaheuristics for solving a constrained statistical modeling problem in bioinformatics; see \cite{cui2024nature, cui2024applications} for more advanced applications of metaheuristics across many different disciplines. In this paper, we focus on one of the recently developed metaheuristics - the BAT algorithm. We first introduce the general idea of optimization algorithms, history of optimization algorithm and metaheuristics in Section 1. In Section 2, we discuss variants of the BAT algorithm, optimal parameter settings of the BAT algorithm, and some applications of the BAT algorithm and its variants to the field of public health. In Section 3, we apply the BAT algorithm to analyze a real dataset and estimate interesting parameters in a biostatistical problem. We conclude in Section 4 and mention some limitations of the BAT algorithms.

\vspace{-0.5cm}
\subsection{Optimization Algorithm}

An optimization algorithm is designed to solve optimization problems. An optimization problem can be boken down into three parts: the objective, function to be optimized, the variables to be optimized and the constraints in the problem. Objectives can be to find the optimal settings of the variables that minimize the cost of some business activity or result in the highest performance of a machine. Such real problems can often be formulated as finding the optimal solution to some mathematical equations. The variables can be the number of employees to be hired, water usage, type of materials to be used or operating hours. Finally, constraints are conditions that our solutions have to satisfy. For example, factories may have to follow certain environmental regulations which may then result in a  limit  on the water usage amount, or the number of employees working at the same time. In addition, complicating the outcome variable, like the price of a product may have to be restricted to a reasonable range.

The mathematical notation of optimization problems can be written as below:
\begin{align}
\text{minimize or maximize }\hspace{0.5cm} &f_i(x) \hspace{0.5cm}\text{for i = 1, 2, ..., I}  \label{eq:model1}\\
\text{under the condition }\hspace{0.5cm} &h_j(x) = c_j \hspace{0.5cm}\text{for j = 1, 2, ..., J} \\
\text{and/or }\hspace{0.5cm} &g_k(x) < d_k \hspace{0.5cm}\text{for k = 1, 2, ..., K} 
\end{align}
where $f_i(x)$ are the objective functions, $h_j(x)$ are the equality constraints and $g_k(x)$ are the inequality constraints. There are n variables or more specifically, decision variables, to be optimized and they are represented by a $n \times 1$ vector $x=(x_1, x_2, ..., x_n)$. All the functions $f_i$, $h_j$ and $g_k$ are given and $c_1, c_2,..., c_J$, $d_1,...,d_K$ are user-specified constants. As with all real world problems, the first goal is to transform each of them into some objective functions similar to $f_i(x)$, along with appropriate constraints $h_j$ or $g_k$ and maximize or minimize the objective function.

\cite{yang2018mathematical} also suggests that a general form of an algorithm is
\begin{align}
x(t+1) = A(x(t), x_*, p, \epsilon),
\end{align}
where $x(t)$ is the solution at iteration t, $x_*$ is the historical best solution and $p = (p_1, ..., p_K)$ are these tuning parameters that depends on the algorithm (Yang, 2017). He also suggests that some degree of randomness could be added to the algorithm by introducing a vector of randomness $\epsilon$(Yang, 2010).  

The history of optimization algorithm seems to originate from a paper by Pierre de Fermat in 1646 \cite{du2016search}. He proposed a calculus-based method to find the optimum by taking the derivative of a function and set it equal to zero. Joseph-Louis Lagrange also presented an problem that optimizes the total distance between three points in a two dimensional space on the same paper. Since then, many classical optimization methods have been developed, such as the Newton-Raphsons method and gradient descent method. A new era of optimization was started with the initial development of linear programming, also called linear optimization (Leonid Kantorovich, 1939). George Dantzig was the first to propose the simplex algorithm to solve linear optimization problem in 1947 \cite{du2016search}. Many branches of optimization emerged after this discovery, including nonlinear optimization, constrained optimization, combinatorial optimization, global optimization and convex optimization.

There are many ways to classify optimization algorithms based on the different aspects of the problem and algorithm. For example, optimization algorithms can be classified into two groups: deterministic algorithms and stochastic algorithms. Deterministic algorithms always produce the same result given the same set of parameters and initial values. This type of algorithms are often gradient-based and are most efficient in problems with one global optimum \cite{ryu2022large, cui2024nature}. Hill-climbing, which is gradient-based algorithm, is an example of this type. Stochastic algorithms add randomness to the algorithm, so it will produce different results even if we use the same set of parameters and initial values. Many nature-inspired metaheuristics are based on swarm intelligence, and swarm intelligence algorithms are mostly stochastic algorithms. For example, particle swarm intelligence which simulates the movement of a swarm of bird or a school of fish looking for an optimum is a type of such optimization algorithm. BAT algorithm is also a metahuristic that utilizes swarm intelligence by mimicking the searching activity of groups of microbat that use echolocation to find their prey. Another common classification is based on the solution, the details are discussed in the next section.

Although extensive efforts have been put into the development of optimization algorithms, there is no algorithm that is able to guarantee solving all types of optimization problems efficiently. For example, traditional gradient based algorithms are generally efficient for optimizing smooth nonlinear functions but they encounter difficulties when the functions are non-differentiable or multimodal. Further, most of them focus on local search, so there is no guarantee that they will find the global optimum \cite{yang2018mathematical}. \cite{wolpert1997no} proposed the "no free lunch" theorem that claims there is no algorithm that will perform  better than all other algorithms when the problems are averaged across all possible classes of problems. However, real-world problems tend to have certain characteristics and it is possible to identify an algorithm that could outperform others for this specific class of problems. 

\vspace{-0.5cm}
\subsection{Heuristics and Metaheuristics}

Another way to categorize optimization algorithm is by the computational technique and provability of the solution, including exact algorithm and heuristics \cite{cui2024nature}. Exact algorithms, such as dynamic programming, guarantee an optimal solution and the solution is provable. This type of algorithm are known to be resource intensive, and resources are limited in real-world problems. The resources needed to solve the problem increases as the computational complexity grows, such as when we have NP-hard problems. In this case, we may need to trade optimality with computation time. Therefore, exact algorithms are often abandoned and heuristics are used. Among optimization problems, problems that can be solved in polynomial time are called P-problems, where P stands for polynomial time. Otherwise, problems with solution that cannot be found in polynomial time are called NP-hard problems, where NP stands for non-polynomial time. Most combinatorial optimization problems, such as the subset sum problem and traveling salesman problem, are NP-hard problems \cite{cui2024nature}. Heuristics depends on trial and error, they are usually less time consuming than exact algorithms. Therefore, heuristics can be used to solve those "hard" problems by providing a feasible solution in an reasonable time \cite{sroka2018odds}. However, they do not guarantee optimality and may encounter convergence issues \cite{sroka2018odds}. For example, it may get stuck in some local maximum or minimum solutions. 

The prefix "meta" in metaheuristic optimization algorithms refers to those heuristics with "higher level" or "beyond" heuristics \cite{cui2024nature}. Heuristics are problem-specfic algorithms and metaheuristics is   used in optimization problems when heuristics is unable to solve them within a limited time \cite{boussaid2013survey}. Most of the metahuristics are  motivated by animal behavior or law of physics. They invariably are not based on the gradient or the hessian of the objective function and are virtually assumptions free \cite{boussaid2013survey}. Consequently, metaheuristics is increasingly used across disciplines to tackle high-dimensional hard to solve optimization problems.

Although the idea of metaheuristic has been existed for a long time, this term "meta-heuristic" was not used until 1986 when Fred Glover published an article that introduced the Tabu search \cite{sorensen2017history}. Tabu search is developed to improve the techniques to solve combinartorial optimization problems in artificial intelligence. It is a type of local search method. Local search moves from the current solution to an adjacent solution iteratively using some pre-defined criteria. Tabu search creates several tabu lists of solutions based on the type and direction of recent $m$ moves where $m$ is a parameter, and it prohibits future moves back to those solutions as long as they remain on the lists. By applying this step, tabu search can enhance the efficiency of local search and prevent itself from stucking at a local optimum.

Many metaheuristics emerged in the past 30 years. Accoding to \cite{hussain2019metaheuristic}, there are more than 200 nature-inspired algorithm now and the bulk of their applications are primarily in engineering and operational research. Among the algorithms discussed in their article, 49.62\% of the proposed algorithms have been applied to a real-world problems in the initial article, and 47.71\% of them have been tested on several benchmark functions. Chapter 1 in \cite{cui2024nature} provides a somewhat elaborate list of these algorithms.

\section{BAT algorithm}

Echolocation, also known as biosonar, is a system that is used by some animals to communicate and locate objects. For example, there is a subgroup of bat called microbat that uses echolocation. They can emit ultrasonic wave through their mouth or nose, and it bounces back when the waves encounter an object or obstacle. By comparing the difference of the wavelength, loudness and time lag of the emitted ultrasonic wave and reflected echos, they are able to distinguish between an object and an obstacle, as well as the characteristics of the object and obstacle. BAT algorithm (Yang, 2010) is an metaheuristic optimization algorithm that make use of such feature. Yang (2010) implemented the BAT algorithm and compared it with genetic algorithm and particle swarm optimization using 10 different benchmark functions, such as the Rosenbrok's function, the eggcrate function and the Michalewicz's test function. His results suggested that the (basic) BAT algorithm performs much better than the genetic algorithm and standard particle swarm optimization.

Since the appearance of the BAT algorithm, many variants have been developed to improve its performance. For example, Nakamura et al. (2012) proposed the binary BAT algorithm that hybridized the BAT algorithm with the optimum-path forest classifier to select features in a validation set. Another is by Lin et al., (2012) who proposed the chaotic levy flight BAT algorithm that combined a chaotic sequence and a Levy random process with the BAT algorithm to adjust the parameters in biological system and avoid been stucked at a local optimum. The third variant of the BAT algorithm was proposed by Fister et al. (2015) who modified the BAT algorithm with quaternion representation to improve the convergence rate. These BAT algorithms have been repeatedly shown to be capable of successfully solving many optimization problems with applications ranging from engineering to public health. In the next section, we provide an introduction to basic BAT algorithm, compare it with other metaheuristic algorithms, and review some of its variants and recent applications with some details.

\subsection{The Basic BAT Algorithm}

Suppose we have a total of $n$ bats, and each simulated bat starts to fly at location $x_{i}(0)$ in a $m$-dimensional search space where $i$ is an index for bat $i$, $i=1,.., n$. The speed and location of the simulated bats are updated at every iteration. We denote $t$ the location of the $i^{\text{th}}$ bat at the $t^{\text{th}}$ iteration by $x_{i}(t)$, and its speed by $v_{i}(t)$. Write $x_i(t) = (x_{i1}(t), x_{i2}(t), ..., x_{im}(t))$ and write $v_i(t) = (v_{i1}(t), v_{i2}(t), ..., v_{im}(t))$ where the second subscript denotes the dimension of the location. For example, $x_{ij}(t)$ is at $j^{\text{th}}$ dimension of the location of the $i^{\text{th}}$ bat and the $t^{\text{th}}$ iteration. We also define $x_*(t)$ as the best location found from all of the $n$ bats at iteration $t$. 

The ultrasonic wave emitted by each microbat varies in its wavelength, loudness, rate of emission and time lag. According to Yang (2010), one of the key assumptions about BAT algorithm is that bats are allowed to adjust their wavelength, loudness and the rate of emission over time. Time lag is not included due to the computational complexity at high dimension (Yang, 2010). Since the light speed $c$ is a contant value and $c = f\cdot\lambda$, we can use frequency $f$ to measure wavelength $\lambda$. Under his assumption, the domain of frequency $f_i$ for all bats are restricted to $[f_{\text{min}}, f_{\text{max}}]$. Since the loudness $A$ decreases and the rate of emission $r$ increases when the bat find a prey, the assumption is that they change monotonically as the algorithm iterates. These assumptions imply that:
\begin{align}
A_i(t) \to 0, \hspace{0.2cm}\text{and}\hspace{0.2cm} r_i(t) \to r_i(0) \hspace{0.5cm}\text{as}\hspace{0.5cm} t \to \infty,
\end{align}
where $r_i(0)$ is the upper bound(initial) of the emission rate for the $i^{\text{th}}$ bat. The loudness and rate of emission are updated using the following equations:
\begin{align}
A_i(t) &= \alpha \cdot A_i(t-1)\\
r_i(t) &= r_i(0) \cdot [1 - exp(-\gamma t)]
\end{align}
where $0 < \alpha < 1$ and $\gamma > 0$ are two fixed constants that manage the rate of change.  The velocity and location of the $i^{\text{th}}$ bat at iteration $t$ are updated as follows:
\begin{align}
x_i(t) &= x_i(t-1) + v_i(t)\\
v_i(t) &= v_i(t-1) + (x_i(t-1) - x_*(t)) \cdot f_i\\
\text{and}  \hspace{0.3cm} f_i &= f_{\text{min}} + ( f_{\text{max}} -  f_{\text{min}}) \cdot \beta,
\end{align}
where $\beta$ is a random number generated from the uniform distribution on $(0,1)$. Here, the initial frequency $f_i(0)$ is generated from the uniform distribution on $(f_{\text{min}}, f_{\text{max}})$.

Exploration is designed to look for the global optimum. To improve its exploration ability,, we diversify the solutions and avoid been trapped at a local optimum. Here exploitation means the algorithm searches for a solution locally until a local optimum is reached via a local search. An efficient algorithm needs to find a balance between exploration and exploitation, so that it converge in a reasonable time without getting stuck at local optimum (Lin et al., 2013). In the BAT algorithm, the local search relates to the loudness and is accomplished using the following updating equation on the current best solution $x_{*, \text{ old}}$ at that iteration:
\begin{align}
x_{*, \text{ new}} = x_{*, \text{ old}} + \epsilon \cdot \bar{A(t)}
\end{align}
Here $\epsilon$ is a random number generated from the uniform distribution on $(-1,1)$ and $\bar{A(t)}$ is the average loudness across all bats.

The pseudo code of the BAT algorithm is shown in Algorithm 1. Here, $rand$ is a random number generated from the uniform distribution on $(-1,1)$. The first $if$ statement for implementing a method to improve local search, and the second $if$ statement for updating the loudness and emission rate when a new solution is found by the bats. The program runs iteratively until the largest iteration is met. Alternatively, the BAT codes and several other nature-inspired metaheuristic algorithms   are also available in MATLAB, the R archive or self written   packages, like the indo package or  from online resources like 
$$
https://github.com/Valdecy/Metaheuristic-Bat_Algorithm.
$$
\begin{algorithm}[h]
\DontPrintSemicolon
\SetKwInput{Kw}{Objective function}
\Kw{$f(x)$, where $x = (x_1, x_2, ..., x_m)$}
\SetKwInput{Kw}{Input}
\Kw{Initial position $x_i(0)$, velocity $v_i(0)$, frequency $f_i(0)$, emission rate $r_i(0)$, and loudness $A_i(0)$ for a total of $n$ bats.} 

 Set t = 0, Largest iteration T\;
 \While{t $<$ T}{
  Update location $x_i(t)$, velocity $v_i(t)$ and frequency $f_i(t)$ for each bat using equation (2.4) to (2.6).\;
  \If{$rand > r_i(t)$}{
   Select best location $x_{*, \text{ old}}$ among all previous best locations for all bats\;
   Update the location $x_{*, \text{ new}}$ for the selected location with equation (2.7)
   }
  \If{$rand < A_i(t)$ and $f(x_i(t)) < f(x_{*, \text{ new}})$}{
  Accept the new solution\;
  Adjust the loudness $A_i(t)$ and rate of emission $r_i(t)$ using equation (2.2) and (2.3)
  }
  Select the best solution $x_*$\;
  $t \leftarrow t + 1$\;
 }
\SetKwInput{Kw}{Output}
\Kw{Best solution $x_*$}
 \caption{Basic BAT algorithm\label{IR}}
\end{algorithm}

\subsection{Parameter Setting of the BAT Algorithm}

The performance of the BAT algorithm depends on the initialization of parameters; they include the number of bats $n$, loudness $A_i(0)$, rate of emission $r_i(0)$, loudness tunning parameter $\alpha$, emission rate tunning parameter $\gamma$ and the distribution of the frequency $f_i$. In the initial article by Yang (2010), he set $\alpha = \gamma = 0.9$, $n$ = $25$ to $50$, uniform distribution on $(0,1)$ and tested its performance using $5$ test functions. To find the optimal setting of algorithm parameters, Xue et al. (2015) did a further investigation using 28 benchmark functions from  the CEC-2013 test suits. In this study, 25 sets of parameters was tested on each benchmark function $51$ times with $300,000$ iterations. The Friedman and Wilcoxon tests were then performed to determine the overall performance. They concluded that $A_i(0) = r_i(0) = \gamma = 0.9$, $\alpha = 0.95$ and $f_i(0)$ drawn from the uniform distribution on $[0,5]$ is the best parameter setting when holding the total number of bats $n$ constant at $100$. 

\subsection{ Other Algorithms}

Particle swarm optimization (Kennedy and Eberhart, 1995) simulates animal social behaviors, such as herds and flocks. Similar to BAT algorithm, it also searches for optimal solution by adjusting the velocity and location of each individual particles with each iteration. Suppose we have $n$ particles, each particle is at position $x_i$ and its velocity is $v_i$ in a $m$-dimensional space. Let $x_{i,best}(t)$ be local best location for the $i^{\text{th}}$ particle and let $p_*(t)$ be the global best location for all particles at iteration $t$. The updating equation for the particle swarm optimization is:
\begin{align}
v_i(t) &= v_i(t-1) + 2 \cdot rand \cdot (p_*(t-1) - x_i(t-1)) + 2 \cdot rand \cdot (x_{i, \text{best}}(t-1) - x_i(t-1)) \\ 
x_i(t) &= x_i(t-1) + v_i(t)
\end{align}
where $2 \cdot rand$ is a stochastic factor designed to have a mean of 1.

Here, if we replace the stochastic factor in (2.8) with frequency, remove the feature of loudness and rate of emission, and add some information about the current global best solution into our updating equation, we observe the BAT algorithm is somewhat similar to the particle swarm optimization (Yang, 2010).

Harmony search (Geem et al., 2001) simulates the process of searching for the perfect state of harmony in music. The updating equation is:
\begin{align}
x_{\text{new}} = x_{\text{old}} + b_{p} \cdot rand
\end{align}
and the randomization components are:
\begin{align}
P_{\text{random}} = 1 - r_{\text{accept}} \\
P_{\text{pitch}} = r_{\text{accept}} \cdot r_{\text{pa}}
\end{align}
where, $b_p$ is a factor that controls the pitch adjustment, $r_{\text{accept}}$ is the harmony memory accepting rate and $r_{\text{pa}}$ is the pitch adjusting rate that controls the probability of pitch adjustment (Yang, 2009).

Here, if we treat frequency change in BAT algorithm as the pitch adjustment in harmony search, and rate of emission as the harmony memory accepting rate, we also observe that the BAT algorithm can be reduced to a special form of harmony search (Yang, 2010).

\subsection{Variants of BAT algorithm}

\begin{table}[!ht]
\caption {List of Variants of BAT Alogrithms} \label{tab:title}
\centering
\begin{tabular}{clr}
  \hline
 & Algorithm & Reference \\ 
  \hline
1 & Multiobjective BAT algorithm(MOBA) & (Yang, 2012) \\ 
2 & Binary BAT algorithm(BBA) & (Mirjalili et al., 2014) \\ 
3 & Chaotic BAT algorithm & (Gandomi and Yang, 2013) \\ 
4 & Binary BAT algorithmb(BBA) & (Nakamra et al., 2012) \\ 
5 & Hybrid BAT algorithm(HBA) & (Fister Jr. et al., 2013) \\ 
6 & Improved BAT algorithm(IBA) &  (Cai et al., 2016)  \\ 
7 & BAT algorithm with harmony search(BAHS) &  (Wang and Guo, 2013)\\ 
8 & Directional BAT algorithm(DBA) &  (Chakri et al., 2016) \\ 
9 & Enhanced BAT algorithm(EBA) &  (Yılmaz and Küçüksilleb, 2015) \\ 
10 & Improved discrete BAT algorithm(IDBA) &  (Osaba et al., 2016) \\ 
11 & Improved BAT algorithm(IBA) &  (Alihodzic and Tuba, 2014) \\ 
12 & Hybrid self-adaptive BAT algorithm(HSABA) &  (Fister el al., 2013) \\ 
13 & Improved BAT algorithm(IBA) & (Bahmani-Firouzi et al., 2014) \\ 
14 & \shortstack{BAT algorithm based on differential operator \\ and Lévy-flights trajectory (DLBA)} &  (Xie et al., 2013)\\ 
15 & BAT algorithm with mutation (BAM) &  (Wang el al., 2012) \\ 
16 & \shortstack{BAT algorithm with directing triangle \\ -flipping strategy (BA-DTFS)} &  (Cai et al., 2017)\\ 
17 & Improved BAT algorithm (IBA) &  (Yilmaz and Kucuksille, 2013)\\ 
18 & Modified BAT algorithm (MBA) &  (Yilmaz et al., 2014)\\ 
19 & Novel BAT algorithm (NBA) & (Meng et al., 2015) \\ 
20 & Complex-valued BAT algorithm (CBA) &  (Li and Zhou, 2014)\\ 
21 & Chaotic levy flight BAT algorithm(CLFBA) &  (Lin et al., 2012) \\ 
22 & Improved BAT algorithm(IBA) &  (Wang et al., 2015)\\ 
23 & BAT algorithm with gaussian walk(BAGW) &  (Cai et al., 2014) \\ 
24 & \shortstack{Self adaptive modified \\  BAT algorithm(SAMBA)} &  (Khooban and Niknam, 2015) \\ 
25 & \shortstack{Simulated annealing gaussian \\ BAT algorithm (SAGBA)} &  (He et al., 2014)\\ 
   \hline
\end{tabular}
\end{table}

The BAT algorithm has been shown to outperform genetic algorithms and particle swarm optimization on some benchmark functions. However, researchers have identified issues such as poor convergence in high-dimensional problems and low solution diversity. To address these limitations, various modifications and enhancements have been proposed. Table 2 lists 25 BAT variants from Google Scholar, each designed to improve performance in specific ways or for targeted problems. Yang (2012) introduced the \textbf{Multi-Objective BAT Algorithm (MOBA)}, which converts multiple objectives into a weighted sum. MOBA demonstrated fast convergence on four test functions and was successfully applied to the welded beam design problem. For binary optimization, Nakamura et al. (2012) developed the \textbf{Binary BAT Algorithm (BBA)} using a sigmoid function to map velocities to Bernoulli-distributed values, outperforming competitors on three out of five feature selection datasets. Mirjalili et al. (2014) addressed a velocity update issue by introducing a \textbf{V-shaped transfer function}, validated through Wilcoxon's rank-sum test against binary PSO and genetic algorithms. To solve the Traveling Salesman Problem, Osaba et al. (2015) introduced the \textbf{Improved Discrete BAT Algorithm (IDBA)}, which replaces velocity updates with a uniform distribution based on Hamming distance and modifies location updates using 2-opt and 3-opt heuristics.

Other researchers focused on improving convergence speed. Cai et al. (2016) proposed an \textbf{Improved BAT Algorithm (IBA)} with an \textit{optimal forage strategy}, selecting the bat with the highest energy-to-distance ratio instead of the current best solution. They also introduced an \textit{IBA with random disturbance}, replacing the frequency parameter $f_i$ with a randomly generated value when the best solution remains unchanged. Fister et al. (2013) developed a \textbf{Hybrid BAT Algorithm (HBA)} incorporating standard differential evolution to enhance local search. Later, Fister et al. (2015) introduced a \textbf{Quaternion-based BAT Algorithm (QBA)}, using quaternion representation to improve search performance.

The basic BAT algorithm moves all bats toward the best bat, potentially missing nearby optimal solutions. Chakri et al. (2016) introduced the \textbf{Directional BAT Algorithm (DBA)}, which updates locations based on both the best bat and a randomly chosen bat closer to the optimum. They also proposed an adaptive search mechanism, introducing a monotonically decreasing parameter in the local search equation. Additionally, to avoid premature convergence, they revised the loudness and emission rate equations to sustain exploration over more iterations.

\section{Applications of Bat Algorithms in Public Health}

BAT algorithm and its variants have applications in many fields, such as in engineering and artificial intelligence. In this section, we focus on some recent applications of BAT algorithm to tackle public health related issues. 

Diabetes mellitus is one of the most prevalent diseases in the US. It is estimated that 32.2 million people of all ages had diabetes among the overall US population (CDC, 2020). There are three main types of diabetes, namely, type 1, type 2 and gestational diabetes, and around 90\% of those cases are type 2 diabetes (CDC, 2020). Early diagnosis of diabetes is helpful in reducing the risk complications. For example, early detection of type 2 diabetes could reduce the morbidity and mortality of cardiovascular disease (Herman, 2015). Due to the high prevalence of diabetes, a method of detection without performing an actual blood test could help with early diagnosis. Soliman and Elhamd (2015) implemented the chaotic levy-flight BAT algorithm proposed by Lin et al.(2012) and used the Pima Indians Diabetes data set to classify patients with diabetes. By replacing the uniformly generated random variable $\beta$, $\epsilon$ and average loudness $\bar{A}$ with chaotic variables generated from a levy distribution and a chaotic sequence, this method ensures a better balance between local search and global search (Lin et al. 2012). In this study, they compared the chaotic levy-flight BAT algorithm with some of the most widely used classification algorithm, such as artificial neural network and support vector machine. The metric they used was classification accuracy after different types of cross validation, and this algorithm has the highest average accuracy among all the listed algorithms.

BAT algorithm was also applied in the diagnosis of other diseases. Image recognition can be a powerful tool in identifying cerebral microbleed from magnetic resonance imaging(MRI). Lu et al. (2020) used chaotic BAT algorithms in the training process of a convolutional network model to identify cerebral microbleed using brain MRI dataset. They extended the BAT algorithm with gaussian chaotic map to determine the parameters for extreme learning machine. Shrichandran et al. (2017) used a hybrid of glow worm swarm optimization and BAT algorithm in image segmentation of retinal blood vessels. Binu and Selvi (2015) hybridized BAT algorithm with fuzzy classifier to classify lung cancer data. Gálvez et al.(2018) used the basic BAT algorithm in border detection of skin lesions. Kishore et al (2015) utilized BAT algorithm in optimizing the parameters in watermarking medical images sent by unsecured internet.

Metaheuristics are widely used to solve optimization problems in operations research. Travelling salesman problem(TSP) was designed to solve the shortest route of a salesman visiting a list of designated cities and then return to the starting point. The vehicle routing problem(VRP) was first formulated by Dantzig and Ramser (1956) as a generalization of the TSP. Compare with one salesman in TSP, VRP allows multiple trucks with different routes. Osaba et al. (2018) developed a discrete and improved BAT algorithm(DaIBA) based on the improved discrete BAT algorithm (Osaba et al., 2016) to solve a Drug Distribution System with Pharmacological Waste Collection problem. They formulated this problem as a rich vehicle routing problem(RVRP) with three constraints, including pickups and deliveries, asymmetric variable costs, forbidden roads and cost (Osaba et al., 2018). Here, rich means multiple constraints. The aim of this study is to minimize cost during drug distribution and medical waste collection. They compared DaIBA with three other metaheuristics by runing each algorithm 30 time with two statistical tests, the result shows DaIBA performs significantly better than other three algorithms.

\subsection{Parameter Estimation Problems in Models for Analyzing Data from Public Health}
\subsubsection{Generalized Linear Models}
Generalized linear models are widely used in biostatistics to evaluate the association between a outcome variable and a set of predictors. The model parameters are estimated using the maximum likelihood estimation. In many statistical packages, an iterative approach involving the Newton-Raphson method were utilized to solve for the maximum likelihood estimator:
\begin{align}
\hat{\beta_t} = \hat{\beta_{t-1}} - I^{-1}(\hat{\beta_{t-1}}) \cdot U(\hat{\beta_{t-1}})
\end{align}
where $I$ and $U$ are the fisher information and score. The information and score will be re-evaluated at each iteration which is computationally expensive at high dimension. Here, we present an example where we want to find the maximum likelihood estimator of the model parameters using the BAT algorithm and compare our results with a previous study by Sroka and Nagaraja (2018) using another method. They argued that the odds ratio estimated from count data directly is more precise than from the dichotomized count data. Specifically, they compared the results from a logistic model with that from a geometric model for estimating the odds ratio estimates in a medical study.

For a count data $Y = (y_1, y_2, ..., y_n)$, we assume the Geometric($\pi_i$) distribution where $i$ is a index of subject and $\pi$ is the prob of a response from subject $i$ after $i$ attempts. The link function we used is log link $g(\mu) = log(\mu) = log[(1-\pi)/\pi]$. Re-writting, we have:
\begin{align}
\pi_i = \frac{1}{1+exp(x_i^T \beta)}
\end{align}
where $x_i$ is a vector of predictors for subject $i$. The likelihood function of $Y$ is:
\begin{align}
L(\beta|Y) = \prod_{i=1}^{n} (1 - \pi_i)^{y_i} \cdot \pi_i
\end{align}
Using (4.2), we can write the log likelihood function (4.3) of $\beta$ as:
\begin{align}
logL(\beta|Y) = \sum_{i=1}^{n} \{y_i log[\frac{exp(x_i^T \beta)}{1+exp(x_i^T \beta)}] - log[1+exp(x_i^T \beta)]\}
\end{align}
Here, the log likelihood function $logL(\beta|Y)$ becomes an objective function with a $k$ dimensional decision variable $\beta$. The covariance matrix:
\begin{align} 
cov(\hat{\beta}) = I^{-1}(\hat{\beta}) &= (X^T W X)^{-1} \notag\\
&= \left(X^T\begin{bmatrix}
    1-\pi_1 & &0 \\
    & \ddots & \\
    0& & 1 - \pi_n
  \end{bmatrix} X\right)^{{-1}},
\end{align}
where $W$ is a $n \times n $ diagonal matrix with $w_{ii} = (d\mu_i/\eta_i)^2 / Var(y_i)$ and $\eta_i = g(\mu_i)$ is the link.

For the dichotomized data $Z = (z_1, z_2, ..., z_n)$, we assume the Bernoulli($\pi_i$) distribution with probability mass function $P(Z_i = z_i) = \pi_i^{z_i} (1-\pi_i)^{1-z_i}$. The link function is $g(\mu) = logit(\mu) = logit(\pi)$. Similar to the previous steps for geometric model from (3.2) to (3.5), we could find the objective function, i.e. the log likelihood function and covariance matrix for the logistic model:
\begin{align}
logL(\beta|Z) = \sum_{i=1}^{n} \{z_i log[\frac{exp(x_i^T \beta)}{1+exp(x_i^T \beta)}] + (1-z_i)log[1+exp(x_i^T \beta)]\}.
\end{align}

The data set we used came from the German Socio-Economic Panel (SOEP). To re-produce the result in Sroka and Nagaraja (2018) obtained fom using a regular estimation method (Tables 4 and 5), a subset of 7 predictors were selected, including an indicator variable for post-reform, an indicator variable for bad health, education group, age group and log income. The data is  in the COUNT package in R and data analysis was done in MATLAB.

We compared the estimate of odds ratio and its 95\% confidence interval from Sroka and Nagaraja (2018) with our results from using BAT algorithm. The parameter setting for the BAT algorithm is the same as the optimal setting from Xue et al. (2015), with 20 bats and 2000 iterations. Table 3 displays the fitted regression coefficients and standard deviation of the two linear models and, Table 4 and Table 5 report their corresponding estimated odds ratios and confidence intervals from the two methods.

\begin{table}[ht]
\caption {Estimation of Model Coefficient Using BAT algorithm} \label{tab:title}
\centering
\begin{tabular}{rrrrrr}
  \hline
& Geometric & & & Logistic & \\\cmidrule{2-3}  \cmidrule{5-6}
Predictors & Estimate & Std && Estimate & Std \\ 
  \hline
Intercept & -0.293 & 0.546 && -1.431 & 1.006 \\ 
Post-reform & -0.138 & 0.051 && -0.192 & 0.094 \\ 
Bad health & 1.143 & 0.074 && 1.196 & 0.196 \\ 
Education(10.5 - 12 yrs) & 0.113 & 0.059 && -0.096 & 0.110 \\ 
Education(7 - 10 yrs) & 0.031 & 0.071 && -0.254 & 0.129 \\ 
Age(40 - 49 yrs) & 0.048 & 0.063 && -0.083 & 0.116 \\ 
Age(50 - 60 yrs)& 0.188 & 0.072 && -0.021 & 0.137 \\ 
Log income & 0.128 & 0.070 && 0.312 & 0.129 \\ 
   \hline
\end{tabular}
\end{table}

From Tables 4 and 5, we observe that the odds ratio estimates using BAT algorithm are very close to the estimated using regular methods. It also converges early at around 1000 iterations. The confidence interval of the estimates from using BAT algorithm are generally shorter than that from using the regular method which could be a potential advantage of using BAT algorithm.

\begin{table*}[!ht]
\caption {Estimation of Odds Ratio and 95\% Confidence Interval From Geometric Model} \label{tab:title}
\centering
\begin{tabular}{rrrrrr}
  \hline
& BA & & & Regular & \\\cmidrule{2-3}  \cmidrule{5-6}
 & OR & Interval && OR & Interval \\ 
  \hline
Post-reform  & 0.87 & (0.79, 0.96) && 0.87 & (0.79, 0.96)\\ 
Bad health & 3.14 & (2.71, 3.63) && 3.13 & (2.71, 3.63)\\ 
Education(10.5 - 12 yrs) & 1.12 & (1.00, 1.26) && 1.09 &(0.95, 1.24)\\ 
Education(7 - 10 yrs) & 1.03 & (0.90, 1.18) && 0.97 &(0.84, 1.11)\\ 
Age(40 - 49 yrs)& 1.05 & (0.93, 1.19) && 1.05 &(0.93,1.19)\\ 
Age(50 - 60 yrs) & 1.21 & (1.05, 1.39) &&1.21 &(1.05, 1.39)\\ 
Log income & 1.14 & (0.99, 1.30) &&1.13 &(0.99, 1.30)\\ 
   \hline
\end{tabular}
\end{table*}

\begin{table*}[!ht]
\caption {Estimation of Odds Ratio and 95\% Confidence Interval From Logistic Model} \label{tab:title}
\centering
\begin{tabular}{rrrrrr}
  \hline
& BA & & & Regular & \\\cmidrule{2-3}  \cmidrule{5-6}
 & OR & Interval && OR & Interval \\ 
  \hline
Post-reform  & 0.83 & (0.69, 0.99) && 0.82 & (0.68, 0.99)\\ 
Bad health & 3.31 & (2.25, 4.86) && 3.28 & (2.24, 4.82)\\ 
Education(10.5 - 12 yrs) & 0.91 & (0.73, 1.13) && 1.19 &(0.94, 1.51)\\ 
Education(7 - 10 yrs) & 0.78 & (0.60, 1.00) && 1.32 &(1.03, 1.70)\\ 
Age(40 - 49 yrs)& 0.92 & (0.73, 1.15) && 0.92 &(0.73, 1.16)\\ 
Age(50 - 60 yrs) & 0.98 & (0.75, 1.28) &&0.99 &(0.76, 1.29)\\ 
Log income & 1.37 & (1.06, 1.76) &&1.26 &(0.98, 1.62)\\ 
   \hline
\end{tabular}
\end{table*}

We further tested the BAT algorithm on a different data sets using the Poisson model with a log link. The data set is obtained from the COUNT package in R. This data set has 6 variables and 1959 observations. The main outcome $Y$ is the length of hospital stay and the objective function is
\begin{align}
logL(\beta|Z) = \sum_{i=1}^{n} \{y_i \cdot (\beta^T x_i)  - \exp(\beta^T x_i) - \log(y_i!)\},
\end{align}
and the covariance matrix of are estimated using:
\begin{align} 
cov(\hat{\beta}) =  \left(X^T\begin{bmatrix}
    \lambda_1 & &0 \\
    & \ddots & \\
    0& & \lambda_n
  \end{bmatrix} X\right)^{{-1}}.
\end{align}

Table 6 is the estimated model parameters from the BAT algorithm and using regular method. Using the same parameters setting with $100$ bats, the algorithm converges between 1300 and 1400 iterations and results are similar to that from the regular method.  

\begin{table*}[!ht]
\caption {Estimation of Model Parameters and Std From Poisson Model} \label{tab:title}
\centering
\begin{tabular}{rrrrrr}
  \hline
& BA & & & Regular & \\\cmidrule{2-3}  \cmidrule{5-6}
 & Estimate & Std && Estimate & Std \\ 
  \hline
Intercept  & 0.513 & 0.150 && 0.524 & 0.150\\ 
Systolic blood pressure  & -0.307 & 0.060 && -0.307 & 0.060\\ 
Procedure & 1.130 & 0.019 && 1.130 & 0.019\\ 
Gender & 0.011 & 0.002 && 0.011 &0.002\\ 
Age & -0.106 & 0.018 && -0.106 &0.018\\ 
Type & 0.189 & 0.017 && 0.189 &0.017\\ 
   \hline
\end{tabular}
\end{table*}

\subsubsection{Find MLE for Log-binomial Model}

The Log-binomial model is sometimes used to estimate the relative risk ratio directly, controlling for confounders  when the outcome is not rare (Blizzard et al., 2006). However, one notorious problem using log-binomial model in practice is that the iterative methods provided in statistical software usually fail to find the MLE or the convergence cannot be attained. For instance, Blizzard et al. (2006) implemented a simulation study to examine the successful rate of fitting log-binomial model where 12 data generation designs were proposed. Their results showed that only around 20\% of well-designed simulated samples had no problem fitting a log-binomial model. The main reason for the fitting problem is that the optimum may be near the boundary of the constrained space or out of the space, where we require the parameter vector to be admissible.  Standard iterative methods in software packages are likely to update steps that lead to estimates outside the constrained space and result in a crash. 

A solution for better fitting a log-binomial model was proposed by de Andrade et al. (2018), where  the original constrained problem of finding MLE is replaced by a sequence of
penalized unconstrained sub-problems whose solutions converge to the solution of the original
program. To  solve each unconstrained sub-problem, they recommended using derivative-free algorithms such as Nelder-Mead or BFGS. The results show that more than 93\% of their simulations produce the correct parameters for the log-binomial models. 

In this subsection, we show the flexibility of CSO-MA and demonstrate that it also an effective method for finding MLEs for challenging models, like the log-binomial model. Because the objective is to maximize 
the log-likelihood function with subject to the constraint that the parameter vector should be admissible, we make it an unconstrained optimization problem by minimizing the following penalized function:
$$-\text{log-likelihood}+\text{penalty of inadmissibility}.$$

We used all 12 simulated scenarios implemented in Blizzard et al. (2006) and were able to confirm that CSO-MA can correctly and stably find all the MLEs with each one running time less than 1 CPU second. In our package, users are able to estimate parameters for one-factor log-binomial models by  calling the  commands:

\lstinputlisting[]{mlelb.m} 

where $\mathsf{X}$ and $\mathsf{Y}$ stand for input data; $\mathsf{beta}$ is the vector of model parameters. For high-dimensional models, i.e., models that contain multiple factors, our $\mathsf{MATLAB}$ codes can be easily accommodated so that the model parameters can be estimated similarly.  

In the following, we compare BAT with several popular nature-inspired algorithms for finding MLE in log-binomial models using simulated data. The model for log-binomial regression stipulates the outcome   $y_i\sim \text{Binomial}(n_i,p_i)$ and
\begin{equation}
\label{equation: nb regression}
\begin{split}
 \log p_{i} &= \beta_0 + \beta_1x_{i1} +  \beta_2x_{i2} + \cdots + \beta_kx_{ik} \\  
  \end{split}
\end{equation}

Since $p_i$ is within range 0 to 1, $\log p_{i}$ is always non-positive, leading to the linear constraint $\beta_0 + \beta_1x_{i1} +  \beta_2x_{i2} + \cdots + \beta_kx_{ik}\le 0$. Hence, the optimization problem is:
\begin{equation}
\label{equation: nb regression}
\begin{split}
 \min_{p_i}&-\sum_{i=1}^n\log\mathbf{P}(Y_i=y_i|p_i)\\
 \text{s.t. }& \beta_0 + \beta_1x_{i1} +  \beta_2x_{i2} + \cdots + \beta_kx_{ik}\le 0.
\end{split}
\end{equation}

For the simulation study, we have three categories of the log-binomial model as in Williamson (2013), where the maximizer of  (\ref{equation: nb regression}) is on a finite boundary, inside the parameter space or  is in the limit. Table~\ref{tab:data2} contains three data sets for estimating  the two parameters $\beta_0$ and $\beta_1$ in the model with only one covariate $X$   taking on three possible values $\{-1,0,1\}$.  Fig.\ref{fig:data_mle} displays  the contour plots of the negative log likelihood function constructed from each data set and we observe that its maximum is at the boundary,  in the limit of the parameter space and in its interior, respectively,  using data sets from left to right. 

\begin{table}[htbp!]
\parbox{.3\textwidth}{
\centering
\begin{tabular}{ccc} 
 \hline
 & (Y=1) & (Y=0)\\
 \hline (X=-1) & 10 & 8\\
  (X=0) & 18 & 9\\
  (X=1) & 5 & 0 \\
 \hline
\end{tabular}
\caption*{MLE at boundary}
}
\hfill
\parbox{.3\textwidth}{
\centering
\begin{tabular}{ ccc } 
 \hline
 & (Y=1) & (Y=0)\\
 \hline (X=-1) & 0 & 17\\
 (X=0) & 0 & 21\\
  (X=1) & 0 & 12 \\
 \hline
\end{tabular}\label{tab:data2}
\caption*{MLE at infinity}
}
\hfill
\parbox{.3\textwidth}{
\centering
\begin{tabular}{ ccc } 
 \hline
 & (Y=1) & (Y=0)\\
 \hline (X=-1) & 2 & 2\\
  (X=0) & 14 & 3\\
  (X=1) & 2 & 17 \\
 \hline
\end{tabular}
\caption*{MLE at interior}
}
\caption{Three data sets from Williamson (2013).}\label{tab:data2}
\end{table}

\begin{figure}[!h]
\caption{Contour plots of the negative log likelihood functions from the three data sets. Left: MLE at boundary; middle: MLE at infinity; right: MLE at interior.}
\label{fig:data_mle}
\end{figure}

How does BAT perform relative to other metaheuristic algorithms?  We mentioned at the onset that there are many other nature-inspired metaheuristic algorithms and it is good practice not to rely on results from a metaheuristic algorithm since such algorithms do not guarantee convergence to the optimum.  Accordingly, we additionally implement some of CSO competitors   to find the MLEs of parameters in the same log binomial model using each of the three data sets, except that we now restrict values of both $\beta_0$ and $\beta_1$ in the range $[-10, 10]$.   Table~\ref{tab:log_bi} reports the comparison results, where NLL refers to the negative log likelihood. We run each algorithm 100 times with 100 iterations using their default values and report the estimated quantities, along with their means and the standard deviations.

\small
\begin{table}[ht!]
\parbox{.99\linewidth}{
\centering
\begin{tabular}{ ccccccc } 
 \hline
 & CSO-MA& ABC & BAT & CS & GA & PSO \\ 
 \hline NLL & 29.78 (0.011) & 30.28 (0.72) & 29.91 (0.17) & 34.81 (4.21) & 43.50 (12.88) &29.77 (0.0013)  \\ 
 $\beta_0$ & -0.34 (0.0086) & -0.37 (0.069) & -0.34 (0.019) & -0.59 (0.24) & -0.97 (0.57) & -0.34 (0.0029)  \\ 
 $\beta_1$ & 0.34 (0.0090) & 0.29 (0.083) & 0.31 (0.035) & 0.12 (0.31) & -0.19 (0.58) & 0.34 (0.0029)\\
 \hline
\end{tabular}
\caption*{MLE at boundary}
}\\
\hfill
\parbox{.99\linewidth}{
\centering
\begin{tabular}{ lllllll } 
 \hline
 & CSO-MA& ABC & BAT & CS & GA & PSO \\ 
 \hline NLL & 2.25 (2$\times10^{-10}$) & 2.25 ($3\times10^{-10}$) & 2.25 (1$\times 10^{-6}$)& 2.86 ($4\times10^{-4}$) & 3.28 (2$\times10^{-3}$) &2.25 (0)  \\ 
 $\beta_0$ & -10 (0) & -10 (0) & -10 (0) & -9.84 (0.14) & -9.74 (0.31) & -10 (0)  \\ 
 $\beta_1$ & 0.17 (4$\times 10^{-4}$) & 0.17 (2$\times 10^{-4}$) & 0.17 (0.025) & 0.19 (0.50) & 0.22 (0.37) & 0.17 (0)\\
 \hline
\end{tabular}
\caption*{MLE at infinity (NLL $\times 10^{-3}$)}
}
\hfill
\parbox{.99\linewidth}{
\centering
\begin{tabular}{ ccccccc } 
 \hline
& CSO-MA& ABC & BAT & CS & GA & PSO \\ 
 \hline NLL & 24.14 (2.44$\times10^{-4}$) & 24.15 (0.025) & 24.17 (0.069) & 26.41 (1.98) & 27.53 (4.15) &24.14 (0.068)  \\ 
 $\beta_0$ & -0.70 (0.0024) & -0.71 (0.022) & -0.69 (0.020) & -0.88 (0.26) &-1.11 (0.37) & -0.71 (0.038)  \\ 
 $\beta_1$ & -0.47 (0.0023) & -0.47 (0.021) & -0.45 (0.033) & -0.42 (0.29) & -0.66 (0.24) & -0.47 (0.041) \\
 \hline
\end{tabular}
\caption*{MLE at interior}
}
\caption{Minimized negative log likelihood values and estimated parameters. ABC = Artificial Bee Colony Algorithm, BA = Bat Algorithm, CS = Cuckoo Search, GA = Genetic Algorithm, PSO = Particle Swarm Optimization.}\label{tab:log_bi}
\end{table}

The results shown in the table support that CSO-MA is an effective algorithm for solving the optimization problem.  In particular, we observe   that the optimized NLL values from CSO-MA are generally smaller than other NLL values found by the other algorithms and the two parameters are estimated more accurately compared with the larger standard errors from the other algorithms. Genetic algorithm (GA) clearly performs the worst followed by Cuckoo Search (CS) in this particular study, but  this does not imply that they are inferior algorithms at all. They may well excel in tackling other optimization problems but these are just intriguing aspects of metaheuristics.  The take home message is that one should compare results from a metaheuristic algorithm with several other algorithms and have a higher confidence that the generated results are correct or nearly so when other algorithms   also produce similar results.
\section{An Epidemiology Example with Non-convergence Issue of Relative Risk Regression}

\cite{marschner2012relative} proposed a relative risk regression model with latent variables for the analysis of heart attack data from the ASSENT-2 study \cite{van1999single}. They reported that there is a convergence issue with the usual ``\textit{glm}" function in R. As an alternative, they propose a variant of the EM algorithm for parameter estimation \cite{gupta2011theory}. In this section, using the heart attach dataset, we show that Metaheuristics can solve the issue easily and produce comparable or even better parameter estimates compared with their developed approach in terms of likelihood function values.

Consider a group of $m$ patients with common covariates, e.g., they receive a common treatment for heart disease and they all come from the same state in the US. Let $x$ be the vector of covariates and assume that there are $y$  of the $m$ patients died after a certain period.  Using the notation of \cite{marschner2012relative}, the general relative risk regression is a variant of log-binomial model where the link function is specified by logarithm, i.e., the probability that a patient is died after a certain period of time given the covariate vector $x$ is
\begin{align}
    p=P(x;\alpha)&=e^{x^T\alpha}
\end{align}

where $\alpha$ is the regression coefficient. The obvious constraint is that $x^T\alpha$ must be less or equal to $0$ and this is the reason \textit{glm} fails to converge. Given $n$ independent groups of patients and each group with common covariates $x_i,i=1,\cdots,n$, the corresponding estimation problem can be formulated as
\begin{align}\label{eq:log_bin_lik}
    &\min_{\alpha}\ -l(\alpha)\\
    &\text{such that }x_i^T\alpha\le 0,\ i=1,\cdots,n.\nonumber
\end{align}
where $l(\alpha)=\sum_{i=1}^n \left[y_i x_i^T\alpha+(m_i-y_i)\log\left(1-e^{x_i^T\alpha}\right)\right]$ is the likelihood function. Applying the BAT algorithm (the codes is a modification of the well-known package ``\textit{pyswarms}" \cite{miranda2018pyswarms}), the estimation problem converges faster than the logbin-algorithm developed in \cite{marschner2012relative}. The results are given in Table~\ref{tab:logbinomial} and the time cost is given in Table~\ref{tab:time}. All computations were performed on a MacBook Pro (2021) equipped with an Apple M1 Max chip, featuring 64 GB of memory. The operating system used was macOS Sequoia 15.3.1.

\begin{table}[]
    \centering
    \begin{tabular}{lrr}
    \hline
         & BA & logbin \\\hline
         \textbf{Optimal value} $l(\alpha)$&-179.90156&-179.90197 \\\hline
      Intercept  &-4.027&-4.027 \\
      Age (65-75)& 1.104 & 1.104 \\
      Age ($>75$)& 1.927 & 1.926 \\
      Killip class II &0.703&0.704 \\
      Killip class III/IV & 1.377&1.377\\
      Treatment delay ($2-4h$) &0.0590&0.0591 \\
      Treatment delay $>4h$ &0.172&0.171\\
      Region (LA) &0.0756&0.0757 \\
      Region (EE) &0.483&0.482\\\hline
    \end{tabular}
    \caption{Estimated regression coefficients from the log-binomial regression model for 30-day mortality following a heart attack. 
    The coefficients are obtained using two methods: BA  and logbin (Maximum Likelihood Estimation with the EM algorithm). The log-binomial model applies a log link function to estimate relative risks for different covariates. Age, Killip class (a measure of heart failure severity), 
    treatment delay, and region are included as predictors. The similarity of estimates across the two methods confirms the stability and reliability of the EM algorithm 
    in obtaining maximum likelihood estimates for the log-binomial model.}
    \label{tab:logbinomial}
\end{table}

\begin{table}[]
    \centering
    \begin{tabular}{lrr}
    \hline
         & BA & logbin \\\hline
         \textbf{Average time} & 0.355 (0.024) & 0.509 (0.022) \\\hline
    \end{tabular}
    \caption{The average time cost for BA and ``logbin" algorithms after 100 trials. The number in brackets indicate the standard deviation.}
    \label{tab:time}
\end{table}

\section{Estimation and Inference in Markov Renewal Regression Model}

In this subsection, we show BAT algorithm can solve estimating equations and produce M-estimates for  model parameters, that are sometimes more efficient than those from statistical packages and standard optimization procedures such as Newton-Raphson.  \cite{cui2024applications} correctly noted that  metaheuristics is rarely used to solve estimating equations in the statistical community.

In a survival study, the experience of a patient may be modeled as a process with finite states \cite{cui2025tutorial} and modeling is based on transition probabilities among different states. 
The EBMT dataset, provided in the \textit{mstate} R package, originates from the European Society for Blood and Marrow Transplantation. It is a comprehensive dataset used for modeling transitions in multi-state survival analysis. The data tracks the clinical progression of patients undergoing hematopoietic stem cell transplantation, capturing events such as transplantation, relapse, and death. It represents a multi-state framework where patients transition through various predefined states over time. Key variables include patient characteristics (e.g., age and sex), details about transplantation (e.g., donor type and conditioning intensity), and time-to-event data for each transition. This dataset is widely used in research to illustrate methods for analyzing multi-state survival models, particularly in studying the effects of covariates on transition probabilities and cumulative hazards. It serves as a benchmark for demonstrating statistical techniques in survival analysis.

 We utilize the EBMT dataset from the \texttt{mstate} package \cite{de2011mstate} to demonstrate the application of multi-state models in clinical research. This dataset comprises information on 2,279 patients who underwent hematopoietic stem cell transplantation at the European Society for Blood and Marrow Transplantation (EBMT) between 1985 and 1998. Patient trajectories are represented across six distinct states: transplantation (TX), platelet recovery (PLT Recovery), adverse events (Adverse Event), simultaneous recovery and adverse events (Recovered and Adverse Event), relapse while alive (Alive in Relapse), and relapse or death (Relapse/Death).

\begin{figure}
    \centering
    \includegraphics[scale=0.4]{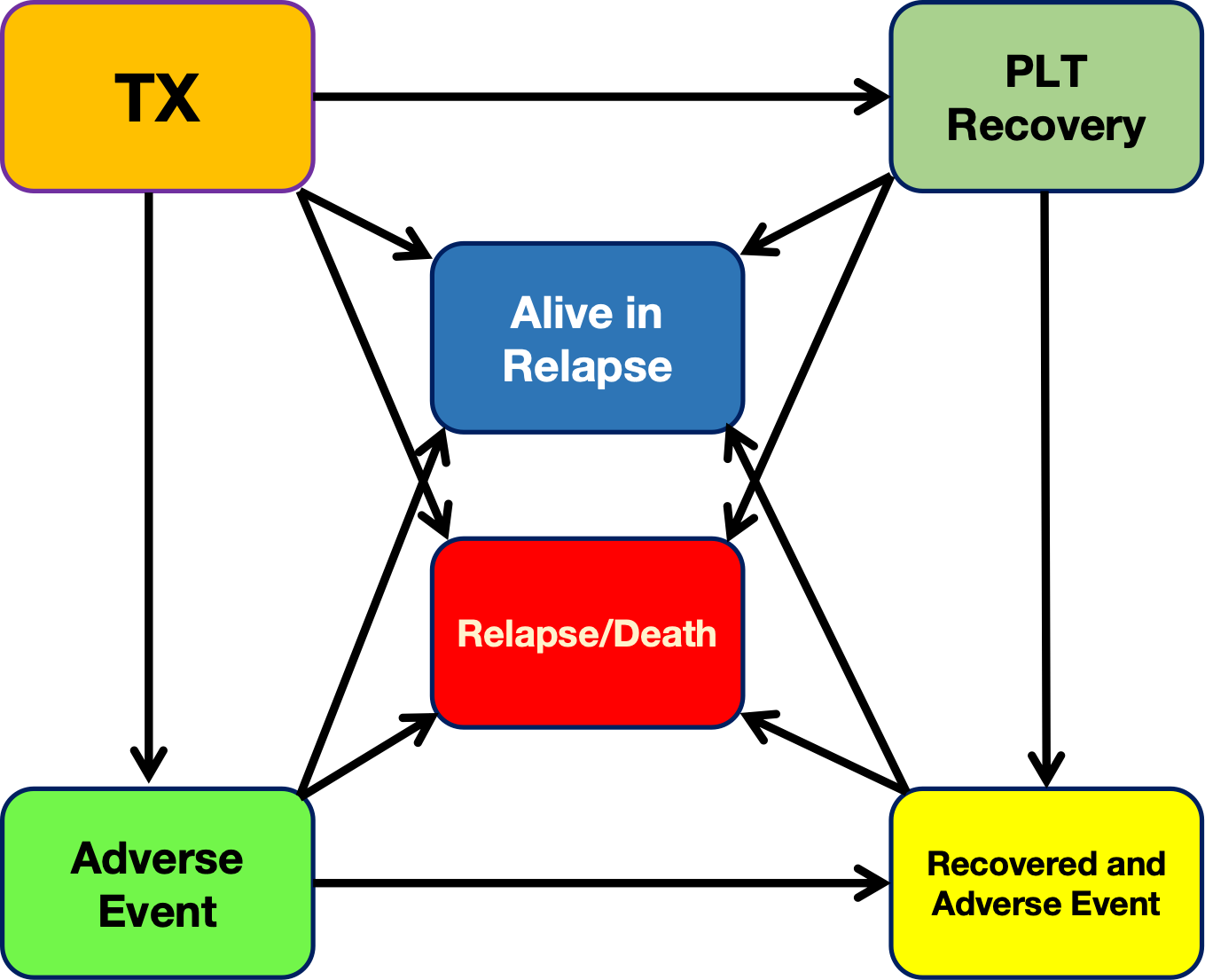}
    \caption{A six-state Markov renewal model for BMT failure. State transition diagram illustrating the possible transitions between clinical states in the EBMT dataset. Temporary states include TX (transplantation), PLT Recovery, Adverse Event, and Recovered and Adverse Event, while absorbing states are Alive in Relapse and Relapse/Death.}
    \label{fig:BMT}
\end{figure}

Figure~\ref{fig:BMT} depicts the possible transitions between these states. All patients begin in the TX state, with potential transitions to PLT Recovery, Adverse Event, or directly to Relapse/Death. From PLT Recovery or Adverse Event, patients may move to other states, such as Alive in Relapse or Relapse/Death. Alive in Relapse is a transient state, while Relapse/Death serves as an absorbing state, marking the conclusion of the clinical trajectory. In addition to state transitions, the dataset includes key covariates such as transplantation year, patient age, prophylaxis use, and donor-recipient gender matching, facilitating covariate-adjusted analyses. Multi-state models applied to this dataset allow researchers to explore the dynamic relationships between recovery, adverse events, relapse, and death. These models provide critical insights into the factors influencing survival trajectories and outcomes for patients undergoing hematopoietic stem cell transplantation.

To model such a process in a mathematically rigorous way, we assume observations on each individual form a Markov renewal process with a finite state, say $\{1,2,\cdots, r\}$ (Dabrowska et al, 1994). That is, we observe a process $(X, T)=\{(X_n, T_n):n\ge 0\}$ where (for simplicity, we do not consider censoring in this subsection), and $0=T_0<T_1<T_2<\cdots$ are calendar times of entrances into the states $X_0, X_1, \cdots, X_n\in\{1,2,\cdots,r\}$. In the EBMT example, $r=6$ and $X_n$ takes values in $\{$TX, PLT Recovery, Adverse Event, Recovered and Adverse Event, Alive in Relapse, Relapse/Death$\}$ and $W_i=T_n-T_{n-1}$ represents the sojourn time staying in the state $X_n$. We also observe a covariate matrix $\mathbf{Z}=\{\mathbf{Z}_{ij}:i,j=1,2,\cdots,r\}$ where each $\mathbf{Z}_{ij}$ itself is a vector. In practice, we assume that the sojourn time $W_n$ given $X_{n-1}=i$ and $\mathbf{Z}$ has survival probability \cite{jacod1975multivariate}
$$\mathbf{P}(W_n>x|X_{n-1}=i, \mathbf{Z})=\exp\left(-\sum_{k=1,k\not=i}^rA_{0,ik}(x)e^{\beta^TZ_{ik}}\right)$$
and the transition probability is ($i\not=j$)
$$\mathbf{P}(X_{n}=j|X_{n-1}=i, W_n)=\frac{\alpha_{0,ij}(W_n)e^{\beta^TZ_{ij}}}{\sum_{k\not=i}\alpha_{0,ik}(W_n)e^{\beta^TZ_{ik}}},$$
where $\beta$ is the parameter of interest, $A_{0,ik}(x)=\int_0^x\alpha_{0,ik}(s)ds$ is the baseline cumulative hazard from state $i$ to state $k$ and $\alpha_{0, ik}(x)$ is the hazard function from state $i$ to state $k$. Suppose we observe $M$ iid individuals and suppose the risk process for an individual is given by $Y_{i}(x)=\sum_{n\ge 1}\mathbb{I}(W_n\ge x, X_{n-1}=i)$. For a fixed $x$, $Y_i(x)$ counts the number of visits to state $i$ with sojourn time more than $x$ for a particular individual. In the six-state model in Figure~\ref{fig:BMT}, since we cannot revisit the states that we have already exited, $Y_i(x)$ is a binary variable.  Then from Dabrowska et al (1994), the partial likelihood for $\beta$ is 
\begin{align}\label{eq:cox}
           L(\beta)= \prod_{n \geq 1} 
\alpha_{X_{n-1}, X_n}(W_n)
\times 
\exp\left[
-\sum_{n \geq 1} \sum_k 
\int_{0}^{W_n} 
\alpha_{X_{n-1}, k}(u) \, du 
\right].
\end{align}
The M-estimates of $\beta$ are obtained by maximizing $L(\beta)$. To apply BAT algorithm to obtain the estimates, we turn the problem into a maximization problem as follows:
\begin{align}\label{eq:U}
\widehat{\beta}=\arg\max_\beta L(\beta).
\end{align}
 \textcolor{black}{Using metaheuristics to creatively solve the system of nonlinear equations, results from our real data example} suggest that the BAT algorithm gives comparable estimates compared with standard packages (Table~\ref{tab:markov_renewal}).

\begin{table}[ht]
    \centering
    \begin{tabular}{l|rrr}
    \hline
        Algorithm & BAT & BFGS & \texttt{coxph} Function \\
    \hline
        Best Log-Likelihood (Subset data) & 2044.688 & 2043.802 & 2043.804 \\
        Best Log-Likelihood (Full data) & 21852.661 & 21852.611 & 21852.620 \\
    \hline
    \end{tabular}
    \caption{Comparison of log-likelihood values achieved by different algorithms for fitting the Markov renewal Cox regression model.}
    \label{tab:markov_renewal}
\end{table}

\section{Parameter Estimation in Hawkes Process Models}
We mentioned earlier that are numerous other nature-inspired metaheuristic algorithms and claimed that they can likewise be used creatively to solve an array  of complex and high-dimensional optimization problems.  Some may perform better than others for specific problems but it is hard to categorize under what circumstances one will outperform another.  In practice, one should run a few such algorithms, and confirm that they produce similar answers to have  higher confidence  that the optimum found is correct or nearly correct. This is because almost all of the metaheuristic algorithms do not have a proof of convergence, even though they seem to work well broadly.

In this section, we demonstrate the potential usage of metaheuristics to Hawkes process models, which has been applied in epidemiology to model the progress of infectious disease such as COVID-19 \cite{rizoiu2018sir, garetto2021time}. \cite{hawkes1971spectra}, Hawkes process is a self-exciting point process that can be applied to model the dynamics of disease progression and is proposed in 1971 \cite{hawkes1971spectra}. Different algorithms have been proposed to simulate a Hawkes process, see e.g., \cite{ogata1981lewis, ogata1998space, daley2003introduction, moller2005perfect, laub2021elements}. Briefly speaking, a Hawkes process is an extension of the non-homogeneous Poisson process such that the intensity increases as more events happen. Mathematically, let $N(t),t\ge 0$ denote the number of infected patients at time $t$, then we assume that the intensity of $N(t)$ satisfies \cite{hawkes1971spectra}
\begin{align}
    \lambda(t)&=\nu+\int_0^tg(t-u)dN(u)=\nu+\sum_{t_i<t}g(t-t_i)
\end{align}
where $\nu>0$ is a constant parameter of a Poisson process, $t_i$'s are jump times of $N(t)$ and $g$ is known as the triggering function. We take it to be exponential, i.e., $g(x)=\alpha\exp(-\beta x)$ where $\beta>\alpha>0$ are hyper-parameters. One can think of $\lambda(t)$ in the following way: at the beginning, the spread of COVID-19 is at a low speed; as more and more patients are infected, the spread of COVID-19 speeds up (the addition terms $g(t-t_i)$ in $\lambda(t)$). This is indeed the case when the COVID-19 pandemic first breakout \cite{fanelli2020analysis, bavel2020using}. Suppose within a pre-specified time range $[0, T]$, we observe $k$ (infection) events with time points $t_1,t_2,\cdots,t_k$. Then the likelihood of the Hawkess process is \cite{ozaki1979maximum}
\begin{align}
L(\nu,\alpha,\beta)=\left(\prod_{i=1}^k\lambda(t_i)\right)\exp\left(-\int_0^T\lambda(t)dt\right).
\end{align}

Different algorithms for simulating $N(t)$ and explicitly calculating $\log L(\nu,\alpha,\beta)$ leads to various computational time. In this section, we adopts the algorithm implemented in the R package \textit{hawkes}. It uses Ogata's algorithm \cite{ogata1981lewis} to simulate both univariate and multivariate Hawkes processes. We compare the BAT algorithm with other metaheuristic algorithms implemented in the R package \textit{metaheuristicOpt} for deriving the MLE of $(\nu,\alpha,\beta)$ \cite{riza2018metaheuristicopt}. The true parameter vector is $(0.2, 0.5, 0.7)$ and we run each algorithm 100 times with 300 iterations each to get reasonable statistical results. The tuning parameter are set to the default values in the \textit{metaheuristicOpt} package and the swarm size is set to 30 for all algorithms.

\begin{figure}
    \centering
    \includegraphics[scale=0.5]{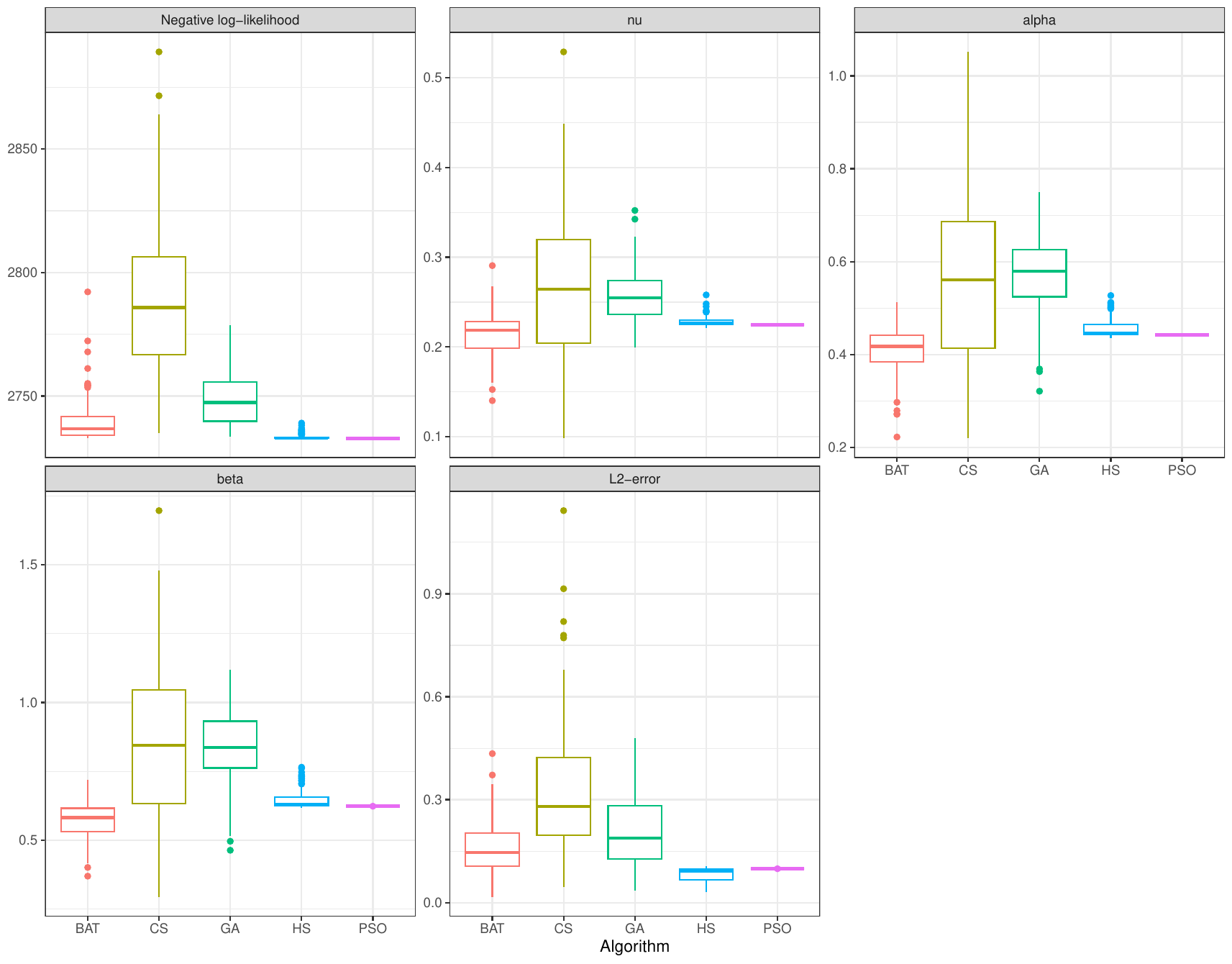}
    \caption{Negative log-likelihood, estimated parameters and $L_2$-error of 100 simulated Hawkes process estimates. BAT = Bat algorithm, CS = Cuckoo search, GA = Genetic algorithm, HS = Harmony search, PSO = Particle swarm optimization.}
    \label{fig:hawkes}
\end{figure}

The average and standard errors of negative log-likelihood, estimated $\nu,\ \alpha$ and $\beta$ and $L_2$-error are reported in Table~\ref{tab:hawkes} and Figure~\ref{fig:hawkes} (the values in brackets are standard error estimates based on 100 runs). Given an estimate $(\widehat{\nu},\widehat{\alpha},\widehat{\beta})$, the $L_2$-error is defined as $\frac{1}{3}\left((\widehat{\nu}-\nu)^2+(\widehat{\alpha}-\alpha)^2+(\widehat{\beta}-\beta)^2\right)$. From the table, we see that PSO performs the best among 5 algorithms in terms of negative log-likelihood values and standard errors. All of PSO standard errors are 0.000 because they are too small to report, i.e., they are smaller than $10^{-10}$. In addition to PSO, Harmony search (HS) algorithm has the smallest $L_2$-error and produces stable parameter estimates. Bat algorithm has an intermediate performance among all 5 algorithms.  Finally, Cuckoo search (CS) and Genetic algorithm (GA) do not produce stable results as other 3 algorithms do.

\begin{table}[!ht]
    \centering
    \begin{tabular}{c|ccccc}
    \hline
    Algorithm & Negative log-likelihood & $\nu$&$\alpha$&$\beta$&$L_2$-error \\
         \hline
BA  &  2739.851 (9.458) & 0.214 (0.026) & 0.408 (0.054) & 0.572 (0.075) & 0.164 (0.087)
\\
        CS & 2788.553 (29.623) & 0.260 (0.086) & 0.563 (0.182) & 0.859 (0.285) & 0.337 (0.201) \\
        GA & 2749.475 (12.023) & 0.257 (0.031) & 0.568 (0.090) & 0.833 (0.144) & 0.207 (0.112) \\
        HS & 2733.447 (1.222) & 0.029 (0.006) & 0.457 (0.022) & 0.648 (0.036) & 0.082 (0.021) \\
        PSO & 2732.831 (0.000) & 0.225 (0.000) & 0.442 (0.000) & 0.624 (0.000) & 0.099 (0.000) \\
         \hline
    \end{tabular}
    \caption{Average and standard errors of negative log-likelihood, estimated $\nu,\alpha$ and $\beta$ and $L_2$-error based on various metaheuristic algorithms. BAT = Bat algorithm, CS = Cuckoo search, GA = Genetic algorithm, HS = Harmony search, PSO = Particle swarm optimization.}
    \label{tab:hawkes}
\end{table}

\section{Conclusion }

This paper reviews one of the popular metaheuristic optimization algorithms based on swarm intelligence called BAT algorithm. We discussed the history of metaheuristics and the inspiration of the BAT algorithm, along with their applications to analyzing data from public health  more efficiently. The detailed mathematical specification and pseudo code were provided. Then, we compared BAT algorithm with other metaheuristics in terms of specification and performance. Some recent developments and applications were provided. Our conclusion is that the BAT algorithm can be widely applied in a great variety of problems and it is easy to implement.

The BAT algorithm and many of its variants have been tested on some sets of pre-defined parameters and they are generally as competitive as current ones or outperform them. As with all metaheuristic algorithms, a more detailed study on the optimal parameter setting is needed to gain better efficiency. Yet the BAT algorithm is just one of many nature-inspired metaheuristic algorithms in the literature to optimize all kinds of complex optimization problems without requiring technical assumptions, like the objective function be differentiable  or even  be explicitly written down. Metaheuristics  offers many opportunities, especially those guided by animal instincts, can help solve public health problems that invariably comprise some large scale optimization problems.  For example, such techniques have been rapidly applied to tackle various aspects of COVID-19 in the last few years.

\bibliography{ref}
\bibliographystyle{apalike}

\end{document}